\documentclass[10pt,twocolumn,letterpaper]{article}

\usepackage{cvpr} 
\makeatletter
\@namedef{ver@everyshi.sty}{}
\makeatother
\usepackage{tikz}
\usepackage{graphicx}
\usepackage{amsmath}
\usepackage{amssymb}
\usepackage{booktabs}

\usepackage{times}
\usepackage{epsfig}
\usepackage{graphicx}
\usepackage{amsmath}
\usepackage{amssymb}
\usepackage{booktabs}
\usepackage{multirow}
\usepackage{multicol}
\usepackage{graphicx}
\usepackage{tikz}
\usepackage{comment}
\usepackage{amsmath,amssymb} % define this before the line numbering.
\usepackage{color,colortbl}

\usepackage[accsupp]{axessibility}  % Improves PDF readability for those with disabilities.
\usepackage{svg}

\usepackage{soul}
\usepackage{mathtools}
\usepackage[linesnumbered,ruled]{algorithm2e}
\usepackage{breqn}
\usepackage{subfiles} 

\SetAlFnt{\small}
\SetAlCapFnt{\small}
\SetAlCapNameFnt{\small}
\SetAlCapHSkip{0pt}
\SetKwInput{KwInput}{Input}                % Set the Input
\SetKwInput{KwOutput}{Output}              % set the Output

\definecolor{Gray}{gray}{0.93}
\definecolor{Gray-lighter}{gray}{0.97}

\usepackage[pagebackref,breaklinks,colorlinks]{hyperref}

\usepackage[capitalize]{cleveref}
\crefname{section}{Sec.}{Secs.}
\Crefname{section}{Section}{Sections}
\Crefname{table}{Table}{Tables}
\crefname{table}{Tab.}{Tabs.}

\def\cvprPaperID{*****} % *** Enter the CVPR Paper ID here
\def\confName{CVPR}
\def\confYear{2022}

\begin{document}

\title{GlobalFlowNet: Video Stabilization using Deep Distilled Global Motion Estimates}

\author{Jerin Geo James \hspace{2em} Devansh Jain \hspace{2em} Ajit Rajwade \\
Indian Institute of Technology Bombay \\
{\tt\small \{jeringeo,devanshdvj,ajitvr\}@cse.iitb.ac.in}
}

\maketitle

\begin{abstract}
Videos shot by laymen using hand-held cameras contain undesirable shaky motion. Estimating the global motion between successive frames, in a manner not influenced by moving objects, is central to many video stabilization techniques, but poses significant challenges. A large body of work uses 2D affine transformations or homography for the global motion. However, in this work, we introduce a more general representation scheme, which adapts any existing optical flow network to ignore the moving objects and obtain a spatially smooth approximation of the global motion between video frames. We achieve this by a knowledge distillation approach, where we first introduce a low pass filter module into the optical flow network to constrain the predicted optical flow to be spatially smooth. This becomes our student network, named as \textsc{GlobalFlowNet}. Then, using the original optical flow network as the teacher network, we train the student network using a robust loss function. Given a trained \textsc{GlobalFlowNet}, we stabilize videos using a two stage process. In the first stage, we correct the instability in affine parameters using a quadratic programming approach constrained by a user-specified cropping limit to control loss of field of view. In the second stage, we stabilize the video further by smoothing global motion parameters, expressed using a small number of discrete cosine transform coefficients. In extensive experiments on a variety of different videos, our technique outperforms state of the art techniques in terms of subjective quality and different quantitative measures of video stability. Additionally, we present a new measure for evaluation of video stabilization based on the flow generated by \textsc{GlobalFlowNet} and argue that it is based on a more general motion model in contrast to the affine motion model on which most existing measures are based. The source code is publicly available at \href{https://github.com/GlobalFlowNet/GlobalFlowNet}{https://github.com/GlobalFlowNet/GlobalFlowNet}
\end{abstract}
% \keywords{video stabilization; teacher-student network; low frequency estimation; optical flow}

\section{Introduction}
\label{sec:intro}
Videos acquired by amateur photographers or lay users from hand-held cameras or mobile phones are subject to a large magnitude of undesirable and discontinuous motion. The process of eliminating or reducing this undesirable motion is called video stabilization. In some setups, the camera can be mounted on stable stands or dollies, but this is infeasible in many commonplace scenarios. Some cameras are equipped with hardware such as gyroscopes for stabilization, but the state of the art in video stabilization still adopts software-based approaches due to the gyroscope's cost, weight and error-pone motion estimation \cite{Milanovic2021, Shi_2022_WACV}. Apart from casual hand-held photography, the need for video stabilization also arises in endoscopy \cite{Karargyris2014}, underwater imaging \cite{Oreifej2011} and aerial photography from drones/helicopters \cite{Lim2019}. Many video stabilization techniques consist of three broad steps: (1) estimation of the motion between consecutive or temporally neighboring frames assuming a suitable motion model, (2) temporal motion smoothing assuming an appropriate motion model for the underlying stable video, and (3) re-targeting or warping of the frames of the unstable video so as to generate a stabilized video. There exists a large body of literature on video stabilization, with differences in the way these three steps are executed. Several of these techniques are summarized below.

\noindent\textbf{Related work (Classical Approaches):} Many traditional techniques assume that the motion between consecutive frames can be approximated using 2D affine transformations or homographies \cite{Matsushita2006,Grundmann2011}, and seek to smooth a sequence of such parameters to render a stabilized video. For computing the parameterized motion, many of these techniques make use of robust point tracking methods \cite{Matsushita2006,Grundmann2011,Liu2013}. However, 2D motion models cannot accurately account for the motion between consecutive video frames for scenes with significant depth variation or significant camera motion. Some methods such as \cite{Liu2013} approximate the motion between consecutive frames by means of patch-wise 2D models or homographies. The method in \cite{Liu2014} performs three tasks in an iterated fashion: determining the smooth global background motion between consecutive frames by detecting moving objects, inpainting the flow in those regions, and smoothing per-pixel optical flow vectors across time. There also exist methods which make use of epipolar geometry \cite{Goldstein2012} or various geometrically motivated subspace constraints \cite{Liu2011}. The latter technique requires fairly long feature tracks which may not be available in many real-world videos. Finally, many techniques which use 3D information have also been proposed, for example methods that use structure from motion \cite{Liu2009}, a depth camera \cite{Sun2012} or a light field camera \cite{Smith2009}.

\noindent\textbf{Related work (Deep Learning Approaches):} Deep neural network (DNN) based approaches for video stabilization have become very popular in recent years. The work in \cite{Yu2019} represents the warp fields using the weights of an unsupervised DNN, which minimizes the sum of two terms: a regularizer that encourages the warp fields to be piece-wise linear, and a fidelity term which minimizes the distance between corresponding pixels in consecutive frames in the stabilized video. This approach, though elegant, must evolve the network weights afresh for every video, and has very high computational cost. The work in \cite{Yu2020} trains a DNN offline on a large video dataset with synthetic unstable motion. In an unsupervised fashion, the weights of the DNN are evolved so as to generate warp fields that (1) have dominant low-frequency content in the Fourier domain, and (2) yield minimal distance between corresponding pixels in consecutive frames of the stabilized video. The work uses frame-to-frame optical flow as initial input and requires a number of pre-processing steps to: (1) identify regions with moving objects from the optical flow fields using a variety of segmentation masks for typical foreground objects obtained from \cite{Zhou2018_ACM}, (2) identify regions of inaccurate optical flow, and (3) inpaint all such regions using the PCA-based approach from \cite{Wulff2015}. The work in \cite{Choi2020} trains two DNNs for performing video stabilization via frame interpolation to smooth the motion between consecutive frames. The $(i-1)^{\textrm{th}}$ and $(i+1)^{\textrm{th}}$ frames are linearly warped mid-way toward each other using the bidirectional optical flow between them. The resulting warped frames are passed through a U-Net \cite{Ronneberger2015} to generate the $i^{\textrm{th}}$ intermediate `stabilized' frame. This interpolation process is carried out iteratively which may accumulate blur. To prevent this, the intermediate stabilized frames are also passed through a Resnet \cite{He2016}. The motion smoothing in this approach is always linear without any adaptation of the smoothing parameters to the motion at different time instants or at different depths. Similar in spirit to \cite{Choi2020}, work in \cite{Liu_2021_ICCV, Xu_2021_ICCV} perform full-frame video stabilization by bringing in border-based frame inpainting. However, the approach in \cite{Liu_2021_ICCV} is computationally expensive. The approach in \cite{Wang2019} trains a Siamese network to generate a warp grid for video stabilization using stable and unstable video pairs from the Deepstab dataset \cite{Xu2018}. Their approach is based purely on color without using any motion parameters and does not perform very well. The approach in \cite{Xu2018} uses spatial transformer networks along with adversarial networks for video stabilization, but suffers from problems due to inadequate training data. The work in \cite{Zhao2020}, which is called PWStableNet, uses a supervised training approach based on a cascade of encoder-decoder units to optimize a combination of criteria such as fidelity w.r.t. the underlying stable video, and various motion and feature-based characteristics. This approach has limitations in terms of training data scarcity and generalizability.

\noindent\textbf{Overview of Proposed Approach:} 
A major contribution of our work is a novel method of estimating the global motion between frames of an unstable video, proposed in Sec.~\ref{sec:theory_global_motion}. Our method involves training a network \textsc{GlobalFlowNet} in a teacher-student fashion in such a way that it imposes a smooth and compact representation for the global motion and is designed to not be influenced by the motion in regions containing moving objects. Our method yields global motion representations that are more general than 2D affine or homography transformation and does not require any salient feature point tracking. Given a pre-trained \textsc{GlobalFlowNet}, we achieve video stabilization using a two-stage process comprising a novel global affine parameter smoothing step (Sec.~\ref{subsec:globalStabilization}) and a novel residual level smoothing step (Sec.~\ref{subsec:smooth_dct}) involving low frequency discrete cosine transform (DCT) coefficients of the residual flow. Both these steps smooth the parameters in the temporal direction. The first step acts as a very useful initial condition, whereas the second step is necessary to significantly improve stabilization performance. This is because it works with a global motion model that despite being very compact, is much more general than just 2D affine or homography. Our overall approach for video stabilization is simple, computationally efficient and interpretable. In extensive experiments (see Sec.~\ref{sec:experiments}), it outperforms state of the art techniques in terms of stability measures. We also propose a new video stabilization measure which uses the low frequency representation from Sec.~\ref{subsec:low_pass_module} to quantify the temporal smoothness of the global motion between successive pairs of frames. Our measure uses a more general motion model than existing measures which largely use affine transformations. 
%As mentioned earlier, the motion vector field that needs to be applied to frames of an unstable video in order to stabilize it, is called a warp field, which is often represented using 2D affine transformations. In our approach, we instead represent such warp fields using linear combinations of low frequency components of a discrete cosine transform (DCT) basis, as shown in Sec.~\ref{subsec:stats_warpfields}. We use this representation in Sec.~\ref{subsec:modified_pwcnet} to robustly estimate the `global motion' between temporally neighboring frames, in a manner not affected by the presence of occlusions or moving objects, by introducing a novel `teacher-student' styled \cite{PLiu2019} modification in the well-known PWC-Net architecture \cite{Sun2018} for optical flow. Then we smooth the sequence of parameters (affine motion parameters as well as DCT coefficients) governing the global motion between pairs of frames in small temporal neighborhoods using assumptions regarding the underlying parameter sequences in a typical stable video (Sec.~\ref{subsec:smooth_affine} and Sec~\ref{subsec:smooth_dct}). 

\section{Global Motion Estimation}
\label{sec:theory_global_motion}
\begin{figure*}
    \centering
    \includegraphics[scale=.38]{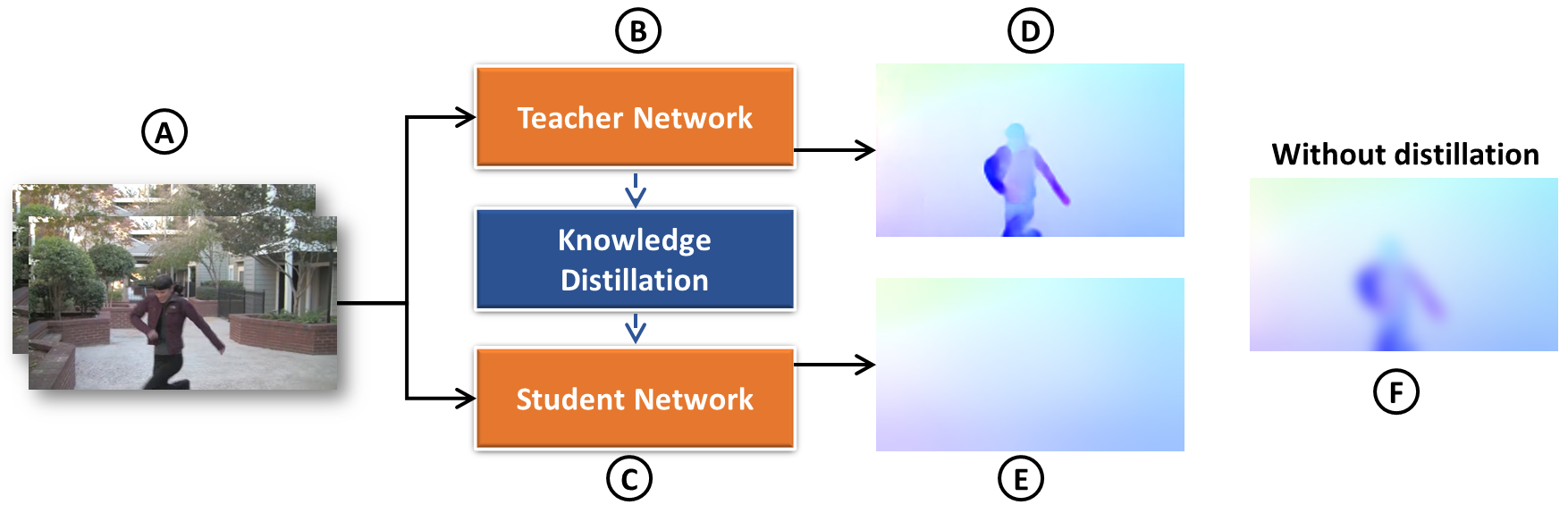}
    \caption{Network architecture for our knowledge distillation approach: a teacher component (B) based on PWC-Net \cite{Sun2018} which produces inter-frame motion estimate (D), and a modified student architecture (C) that obtains smooth global inter-frame motion (E) after training with a robust loss). (F) represents a flow obtained by just low-pass filtering of the flow in (D) without knowledge distillation; it is not part of the network and is shown only for comparison.}
    \label{fig:theory}
\end{figure*}

A key step in video stabilization is the estimation of global motion between consecutive video frames (or temporally nearby video frames), followed by temporal smoothing of the motion parameters. The difference between the original global motion and the global motion in a stabilized video constitutes the \textit{warp field}, which when applied to the unstable frames, stabilizes the video. An ideal global motion or warp field will contain motion discontinuities due to scene depth variations or occlusions. If this warp field is applied directly, it would create holes in the resulting images, requiring non-trivial, potentially error-prone inpainting operations. Hence, many methods approximate the warp fields using continuous motion vector fields or parametrically via 2D affine transformations or homography \cite{Grundmann2011,Matsushita2006}. In this work, we aim to train a network to acquire an inherent ability to produce smooth global motion between consecutive frames in a manner not influenced by independently moving objects in the scene. We achieve this using a knowledge distillation mechanism sketched in Fig.~\ref{fig:theory}. 

\subsection{Knowledge Distillation Approach}
\label{subsec:theory_approach}
Given a standard optical flow estimation network $\mathcal{T}$ which acts as a teacher, we create a student network $\mathcal{S}$, initialized with the weights of $\mathcal{T}$, by introducing a \textit{low-pass filter module} (\textsc{Lpm}). The \textsc{Lpm} is chosen in a manner such that it can represent the global motion between two frames with a high degree of accuracy, but not the motion in regions containing independently moving objects. This is because video stabilization is expected to smooth \emph{only} global inter-frame motion and \emph{not} cause any change to the motion of independently moving objects. Without any further training of $\mathcal{S}$, the optical flow $\boldsymbol{f_S}$ produced by $\mathcal{S}$,  would be a blurred version of the optical flow $\boldsymbol{f_T}$ produced by $\mathcal{T}$, as shown in Fig.~\ref{fig:theory}(F). Such a flow would necessarily contain components of the motion from the independently moving objects leaked into the neighbouring pixels, which would create motion artifacts if used for video stabilization. Instead, we would want $\mathcal{S}$ to mimic the optical flow produced by $\mathcal{T}$ in all regions except the moving objects. This can be achieved by training $\mathcal{S}$ using a robust loss function given as:
\begin{eqnarray}
C_1 = \sum_{l=1}^{N_T} \sum_{i=1}^{N} \mathfrak{R}\left(\|\boldsymbol{f_T}(l,i) - \boldsymbol{f_S}(l,i)\|_2\right), \\
\mathfrak{R}(x; \alpha,c) = \dfrac{|\alpha-2|}{\alpha}\left(\left( \dfrac{\left(x/c\right)^2}{|\alpha -2|} + 1\right)^{\alpha/2} -1\right),
\label{eq:robust_loss_function}
\end{eqnarray}
where $N$ is the number of pixels per image, $N_T$ is the number of image-pairs on which the network is trained, and $i,l$ are indices for pixels and image-pairs respectively. The robust loss $\mathfrak{R}(.)$ that we use here was introduced in \cite{Barron2019}, for which we chose shape parameter $\alpha \triangleq -0.1$ and scale parameter $c \triangleq 0.001$ in our work.

This approach of (1) constraining the dimensionality of $\boldsymbol{f_S}$, and (2) using a robust loss as opposed to a squared error, ensures that the training of $\mathcal{S}$ focuses on global motion and is not influenced by the flow on moving objects. For the student network $\mathcal{S}$, which we henceforth refer to as \textsc{GlobalFlowNet}, we use the well known PWC-Net architecture for optical flow \cite{Sun2018}. For the low pass filter module \textsc{Lpm}, we use low frequency (upto some cutoff frequency) DCT basis vectors. %More details about these components are provided in the following two sub-sections. 
%The details and motivation each of the components viz. $GlobalFlowNet$, $\boldsymbol{LPM}$), and $R$ are explained in the subsequent subsections.
%Note that both \textsc{Lpm} and the robust loss $\mathfrak{R}(.)$ are important and work in tandem. Without \textsc{Lpm}, $\mathcal{S}$ would have just duplicated $\mathcal{T}$. Similarly, without $\mathfrak{R}(.)$, $\boldsymbol{f_S}$ would be a flow which minimizes error with $\boldsymbol{f_T}$ in a least square sense, which would be a smoothed version of the $\boldsymbol{f_T}$.

\subsection{Low Pass Filter Module}
\label{subsec:low_pass_module}
As mentioned before, \textsc{Lpm} should be able to represent global motion with high accuracy and should exhibit poor accuracy in regions with moving objects. With this aim in mind, any low rank off-the-shelf frequency transforms could be used for \textsc{Lpm}. In our method, we choose low frequency components of the Discrete Cosine Transform (DCT). This choice is motivated by our experiments performed on a large video dataset such as \cite{videostabLiu}. For our experiments, we selected pairs of consecutive frames that did not contain independently moving objects. For such frames, we observed that the optical flow, which is the same as the low-frequency global motion, is accurately expressed using \emph{a very small number of DCT coefficients} with frequencies ranging from $(u,v) = (0,0)$ to $(u,v) = (R,R)$ for a cutoff frequency of $R \leq 8$, as can be seen in Figure 1 of the suppl. material at \cite{suppmat}. 
%In the absence of moving objects, the optical flows between consecutive frames are same as global motion between them. We did a statistical analysis on a set of such videos and observed that this global motion can be compactly expressed using few DCT coefficients. This is demonstrated in fig.\ref{fig:stats_warpfields}. 

\subsection{GlobalFlowNet: Global Motion Estimation Network}
\label{subsec:GlobalFlowNet}
The PWC-Net architecture \cite{Sun2018}, on which \textsc{GlobalFlowNet} is based, has three important modules, as illustrated in supplementary material \cite{suppmat}: 
\begin{enumerate}
    \item \textit{Feature extractor}: This module converts the original image into a feature map in each level of refinement.
    \item \textit{Warping layer}: At each level, there is an estimate of the optical flow from previous level. The warping layer warps the features of the target image based on this flow. This layer helps in obtaining the optical flow for small and fast moving objects accurately. 
    \item \textit{Cost volume and context network}: The warped target feature-map is correlated with the source feature-map to obtain a cost volume. Then this cost volume passes through an optical flow estimator and a context network to produce the refined flow for a particular level.
\end{enumerate}
We introduce the following changes to this architecture to obtain our modified (student) network \textsc{GlobalFlowNet}, as illustrated in Fig.~2 of \cite{suppmat}:
\begin{enumerate}
    \item At each level, after the optical flow estimation, we add a \textsc{Lpm} as described in Sec.~\ref{subsec:low_pass_module}. The cutoff frequency for the module is progressively made to increase from the coarse level to the fine level, up to a maximum of 8 (in both directions). 
    \item We also switch off the warping layer from the original PWC-Net, and instead use the motion estimated from the previous layer as an initial condition for the next one. We empirically observed better results as compared to using the warping layer. Moreover, \cite[Table 5e]{Sun2018} shows that excluding this layer does not adversely affect the optical flow accuracy significantly. 
    %The intuition behind this is that, warp layer primarily helps in preserving the motion of small and fast moving objects, which usually is a challenge during coarse to fine optical flow estimation. We also empirically found that switching off this layer helps in our application. Note that switching off  this layer doesn't degenerate the model. The ablation study for warp layer is shown in the original paper in table 5e.
\end{enumerate}
Note that at the time of deployment, only the student network \textsc{GlobalFlowNet} from Fig.~\ref{fig:theory} needs to be used. The teacher network plays a role only during training.

\section{Video Motion Stabilization}
\label{sec:Stabilization}
Once \textsc{GlobalFlowNet} is trained, we use the optical flows produced by it to stabilize a video through a two-stage process. The first stage is a global motion stabilization stage involving correcting for the affine distortions in a novel way, detailed below in Sec.~\ref{subsec:globalStabilization}. This stage corrects a significant amount of global instability. However, since affine transformation is not a good representation for the global motion, this stage leaves behind some spatio-temporal distortions in the video (see also Sec.~\ref{sec:experiments} and many video results in \cite{suppmat}). We correct these residual motion instabilities through a second stage (see Sec.~\ref{subsec:smooth_dct}) involving smoothing of DCT coefficient representation for the remaining global inter-frame motion. 

\subsection{Stage 1: Global Motion Stabilization}
\label{subsec:globalStabilization}
In this stage, we approximate the dense global motion as a coarser affine transformation, and then smooth the sequence of frame-to-frame affine transformation parameters. 

\noindent\textbf{Estimating Affine Transformations:} Affine transformations between consecutive video frames are commonly obtained using salient feature point matching and a robust estimator involving RANSAC. However, this approach is expensive and error-prone if many salient points are concentrated on moving objects or in small regions of the image. Instead, we adopt a novel approach for affine estimation: (1) employing \textsc{GlobalFlowNet} to determine the smooth global motion $\boldsymbol{f_S}$ as detailed in Sec.~\ref{subsec:GlobalFlowNet}, and (2) estimating the $K \triangleq 4$ parameters (rotation angle $r$, translation $t_x,t_y$ and logarithmic scale $s$)  of the (partial) affine transformation directly from the global motion as described in supplementary material \cite{suppmat}. The computation of global motion $\boldsymbol{f_S}$ enables an efficient and fitting of affine motion parameters between adjacent frames without any expensive schemes like RANSAC.

\noindent\textbf{Parameter Sequence Smoothing:}
Consider a parameter sequence $\{\boldsymbol{\alpha_i}\}_{i=1}^{T-1}$ of affine transformations between consecutive pairs of frames $\{(I_i,I_{i+1})\}_{i=1}^{T-1}$ in the unstable video. Here $\boldsymbol{\alpha_i}$ is the vector of $K$ parameters (enlisted earlier) of the affine transformation from frame $I_i$ to $I_{i+1}$, and $T$ is the total number of frames of the unstable video. We need to smooth $\{\boldsymbol{\alpha_i}\}_{i=1}^{T-1}$ to yield a resulting (smoothed) parameter sequence $\{\boldsymbol{\beta_i}\}_{i=1}^{T-1}$, and apply the residual motion sequence $\{\boldsymbol{\gamma_i}\}_{i=1}^{T-1}$, where $\forall i, \boldsymbol{\gamma_i} \triangleq \boldsymbol{\beta_i} - \boldsymbol{\alpha_i}$, to the frames of the unstable video to perform stabilization. However, excessive smoothing of $\{\boldsymbol{\alpha_i}\}_{i=1}^{T-1}$ can lead to a huge loss of field of view. When a video frame is warped using parameters $\boldsymbol{\gamma_i}$ for the purpose of stabilization, some parts of the frame contents will fall outside the usual rectangular canvas and intensity values will remain undefined in some parts. The ratio of the area of the largest inscribed rectangle inside the valid parts of the warped frame to the area of the original rectangular frame is called the \textit{crop ratio} and denoted by $C_R$. We want to ensure that $C_R$ is no less than some user-specified limit $\kappa$. For this, we need to constrain the values $\boldsymbol{\gamma_i}$ using slack parameters $\{\xi_k\}_{k=1}^K$ in such a way that $\forall k \in [K], |\gamma_i(k)| \leq \xi_k$.  
\begin{figure*}
    \centering
    \includegraphics[scale=.32]{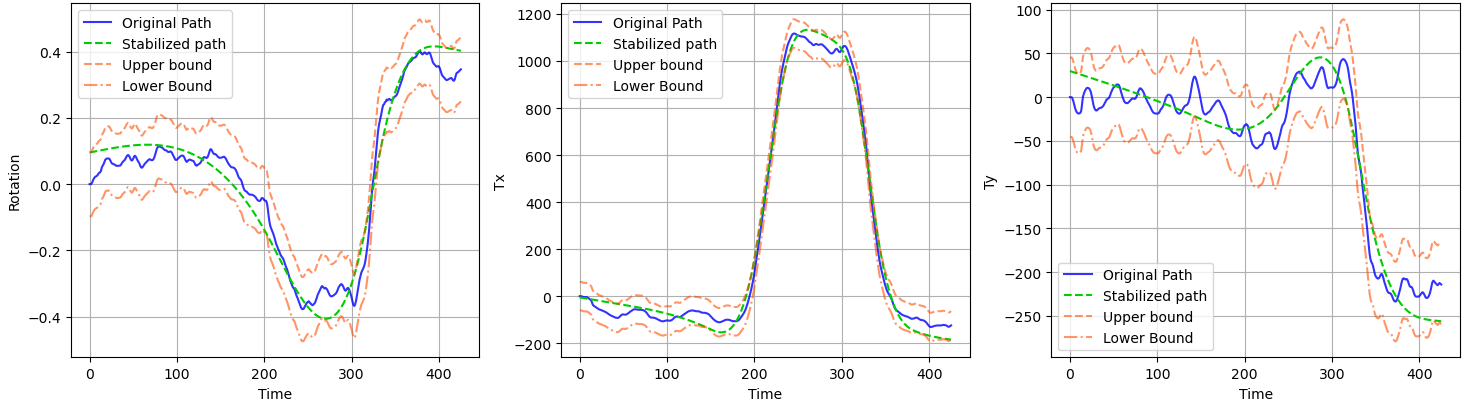}
    \caption{An illustration of affine MPS smoothing using the quadratic programming approach from Eqn.~\ref{eq:smooth} for $\kappa = 0.8$ on a video from the `quick rotation' category in \cite{videostabLiu}.}
    \label{fig:QPath}
\end{figure*}
The cost function we seek to minimize in order to find a smoothed parameter sequence $\{\boldsymbol{\beta_i}\}_{i=1}^{T-1}$ is given as follows:
\begin{eqnarray}
C_2(\{\boldsymbol{\beta_i}\}_{i=1}^{T-1}) \triangleq \sum_{i=2}^{T-1} \|\boldsymbol{\beta_{i+1}}-2\boldsymbol{\beta_i}+\boldsymbol{\beta_{i-1}}\|^2_2 \nonumber \\
\textrm{ such that } \forall i \in [T-1], \forall k \in [K], |\beta_i(k)-\alpha_i(k)| \leq \xi_k.
\label{eq:smooth}
\end{eqnarray}
As per cinematography principles, smoothness of the underlying motion parameter sequence (MPS) is a key feature of stabilized videos, which $C_2(.)$ promotes. Eqn.~\ref{eq:smooth} represents a constrained quadratic programming problem for which fast solvers exist. %In our work, we have considered rotation angle $r$, translation $t_x,_ty$ and logarithmic scale $s$ as $K = 4$ affine parameters.

\noindent\textbf{Choice of slack parameters:} The crop ratio $C_R$ for any frame $I_i$ is a decreasing function of $\{|\gamma_i(k)|\}_{k=1}^K$. Searching for all $\{\xi_k\}_{k=1}^K$ to maintain $C_R \geq \kappa$ is an expensive operation. For simplicity, we express $\forall k \in [K], \xi_k = \lambda_k z$ where $\lambda_k$ is set to be equal to the average local standard deviation of the values from $\{\alpha_i(k)\}_{i=1}^{T-1}$. Given this, we now use binary search to select the maximum value of the parameter $z \in [0,1]$ so that $C_R$ does not fall below $\kappa$. Note that this is a single parameter for the entire video. A set of sample path results obtained by solving Eqn.~\ref{eq:smooth} using $\{\xi_k\}_{k=1}^K$ thus selected, are presented in Fig.~\ref{fig:QPath}.

The main differences between our technique and related MPS smoothing approaches from \cite{Grundmann2011,Liu2013} are that (1) we compute the affine parameters without point matching (see Sec.~\ref{subsec:globalStabilization}), and that (2) we tune $\{\xi_k\}_{k=1}^K$ keeping the crop ratio in mind. On the other hand, the work in \cite{Grundmann2011} puts upper/lower bounds on the values of $\beta_i(k)$, which are less intuitive to specify. The method in \cite[Eqn. 5]{Liu2013} penalizes a weighted combination of the smoothness of $\{\boldsymbol{\beta_i}\}_{i=1}^{T-1}$ and its similarity to $\{\boldsymbol{\alpha_i}\}_{i=1}^{T-1}$, without considering $C_R$.

\subsection{Stage 2: Residual Motion Stabilization}
\label{subsec:smooth_dct}
Consider frame $I_i$ at time instant $i$ in the unstable video. Let us define $\Omega(i;W_R) \triangleq [i-W_R,i+W_R]$ to be a temporal neighborhood of radius $W_R$ around frame $I_i$. Let the estimates of the global motion from $I_i$ to all frames in $\{I_j\}_{j \in \Omega(i;W_R)}$ as produced by \textsc{GlobalFlowNet} be denoted by $\{\boldsymbol{f_S}^{(i,j)}\}_{j \in \Omega(i;W_R)}$. The local sequence $\{\boldsymbol{f_S}^{(i,j)}\}_{j \in \Omega(i;W_R)}$ will be temporally smooth in a stable video, since by design it contains no contribution from independently moving objects (see Sec.~\ref{subsec:GlobalFlowNet}). Therefore, in order to stabilize the given video, we do the following: (1) We extract the global motion from $I_i$ to $\{I_j\}_{j \in \Omega(i;W_R)}$ using \textsc{GlobalFlowNet} from Sec.~\ref{subsec:GlobalFlowNet}, and (2) Apply temporal smoothing filters to smooth the low-frequency DCT coefficients  $\{\boldsymbol{\theta}^{(i,j)}\}_{j \in \Omega(i;W_R)}$ representing the global motion sequence $\{\boldsymbol{f_S}^{(i,j)}\}_{j \in \Omega(i;W_R)}$. 

For (2), we could have followed the quadratic programming strategy from Sec.~\ref{subsec:globalStabilization}. However we observed that a bilateral filter \cite{Tomasi1998} of the following form with temporal smoothing parameter $\sigma_{t} \triangleq W_R/3$ and range smoothing parameter $\sigma_p \triangleq 0.1$ (for intensity values in [0,1]) yielded us good results:
\begin{equation}
\boldsymbol{\widetilde{\theta}}^{(i)} = \dfrac{\sum_{j \in \Omega(i;W_R)} e^{-(i-j)^2/2\sigma^2_t} e^{-\|I_i(\boldsymbol{\Psi\theta}^{(i,j)})-I_j\|^2/2N\sigma^2_p} \boldsymbol{\theta}^{(i,j)}}{\sum_{j \in \Omega(i;W_R)} e^{-(i-j)^2/2\sigma^2_t} e^{-\|I_i(\boldsymbol{\Psi\theta}^{(i,j)})-I_j\|^2/2N\sigma^2_p}},
\end{equation}
where $I_i(\boldsymbol{\Psi\theta}^{(i,j)})$ denotes the image $I_i$ warped by the motion vector field $\boldsymbol{\Psi \theta}^{(i,j)}$ towards image $I_j$. Given the sequence of smoothed DCT coefficients $\{\boldsymbol{\widetilde{\theta}}^{(i)}\}_{i=1}^{T-1}$, the corresponding smoothed global motion estimates $\{\boldsymbol{\Psi\widetilde{\theta}}^{(i)}\}_{i=1}^{T-1}$, where $\boldsymbol{\Psi}$ represents the 2D-DCT, are used to warp the frames $\{\widetilde{I}_i\}_{i=1}^{T-1}$ to generate the stabilized video frame with suitable cropping/resizing. In practice, better results were observed by not smoothing the sequence of zero-frequency DCT coefficients (i.e. DC), but performing smoothing on DCT coefficients of other frequencies. This is because the DC coefficients represent translational motion, and hence that sequence is already smooth due to the procedure in Sec.~\ref{subsec:globalStabilization}.

Our strategy here is a generalization of the affine MPS smoothing technique from \cite{Matsushita2006} to smoothing DCT coefficient sequences. However, it is easy to compose different frame-to-frame affine transformations by iterated matrix multiplication (or parameter addition) to compute the affine motion from frame $I_i$ to its neighbor $I_j$. This is not possible using the DCT coefficient representation. Hence we \emph{separately} estimate the motion from $I_i$ to every member of $\{I_j\}_{j \in \Omega(i;W_R)}$ using \textsc{GlobalFlowNet}. 

\subsection{Summary of Approach and Discussion}
\label{sec:summary_approach}
The exact sequence of steps for implementing our video stabilization approach are summarized in Alg.~\ref{algo:videostab}.

We note that the smoothness of warp fields has been earlier exploited in \cite{Yu2019} based on piece-wise linear approximations, and in \cite{Yu2020} using the Discrete Fourier Transform (DFT). However, we have observed better compactness using the DCT, which is in line with basic principles of image and signal processing \cite[Fig. 8.25, Sec. 8.2.8]{GonzalezWoods} -- see also Fig.1 of \cite{suppmat}.  More importantly, our approach does not require elaborate pre-processing based on pre-trained segmentation masks to identify the foreground, any inpainting of the optical flow in occluded regions or any additional post-processing step on the output of the algorithm, unlike the approaches in \cite{Yu2019,Yu2020}. Therefore, our approach is simpler to implement. 

Our paper presents a novel approach to global motion estimation, including affine transformation estimation, which does not use point tracking. The affine transformation estimates in steps 2--4 of Alg.~\ref{algo:videostab}, bring about a fair degree of stabilization to the original video. However as will be shown later in Sec.~\ref{sec:experiments} and in the accompanying video outputs in \cite{suppmat}, they still retain a lot of wobble artifacts as well as geometric distortion. Due to this, the subsequent steps 5--7 for residual motion smoothing via DCT coefficients are also very important. The main reason for this is that the DCT-based global motion estimates form a compact but more general motion model than just 2D affine transformations. The approach in \cite{Liu2014} also attempts to find global motion that is more general than 2D affine or homography transformations. However it adopts an iterative approach to detect moving objects and inpaint the flow in those regions, in conjunction with smoothing of the motion representation. Besides being iterative, their approach requires the selection of many parameters that may vary across iterations. Our approach here is much simpler to implement and does not require iterated feedback from the intermediate stabilized video.

\begin{algorithm}[!t]
\DontPrintSemicolon
    \KwInput{Input unstable video $\{I_i\}_{i=1}^T$; desirable crop ratio limit $\kappa$; smoothing window radius $W_R$.}
    \KwOutput{Output stabilized video $\{J_i\}_{i=1}^T$.}

    Obtain the global motion estimates $\boldsymbol{f_S}^{(i,i+1)}$ from frame $I_i$ to $I_{i+1}$ for each $i \in [T-1]$ using the pre-trained \textsc{GlobalFlowNet} from Sec.~\ref{subsec:GlobalFlowNet}.\;
    For every $i \in [T-1]$, use a robust method to fit affine transformation parameters $\boldsymbol{\alpha_i}$ to $\boldsymbol{f_S}^{(i,i+1)}$ (Sec.~\ref{subsec:globalStabilization}).\;
    Smooth the sequence $\{\boldsymbol{\alpha_i}\}_{i=1}^{T-1}$ to obtain the sequence $\{\boldsymbol{\beta_i}\}_{i=1}^{T-1}$ using the method in Sec.~\ref{subsec:globalStabilization}.\;
    For each $i \in [T-1]$, warp the frame $I_i$ with suitable cropping to obtain an intermediate stabilized frame $\widetilde{I}_i$.\;
    For every $i \in [T-1], j \in \Omega(i;W_R)$, determine estimates $\boldsymbol{f_S}^{(i,j)}$ of the global motion from $\widetilde{I}_i$ to $\widetilde{I}_{j}$.\;
    For every $i \in [T-1], j \in \Omega(i;W_R)$, determine the DCT coefficients $\{\boldsymbol{\theta}^{(i,j)}\}$ of the global motion estimates from the earlier step. Smooth the temporal sequence of DCT coefficients using a window width $W_R$ in the method in Sec.~\ref{subsec:smooth_dct} yielding $\{\widetilde{\boldsymbol{\theta}}^{(i)}\}_{i=1}^{T-1}$\;
    Obtain the stabilized video by warping the frames $\{\widetilde{I}_i\}_{i=1}^{T-1}$ using the flow reconstructed from the smoothed DCT coefficients (with suitable cropping).
\caption{Video Stabilization Algorithm}
\label{algo:videostab}
\end{algorithm}

\section{Experiments}
\begin{table*}[]
\centering
\resizebox{\textwidth}{!}{%
\begin{tabular}{|clcccccccc|}
\hline
\rowcolor{Gray}
\multicolumn{1}{|l|}{\textbf{Dataset}} &
  \multicolumn{1}{l|}{\textbf{Category}}  & \multicolumn{1}{c|}{\textbf{\begin{tabular}[c]{@{}c@{}}Original\\ \end{tabular}}} & \multicolumn{1}{c|}{\textbf{\begin{tabular}[c]{@{}c@{}}\textsf{BCP}\\ \cite{Liu2013}\end{tabular}}} & \multicolumn{1}{c|}{\textbf{\begin{tabular}[c]{@{}c@{}}\textsf{LVS}\\ \cite{Yu2020}\end{tabular}}} & \multicolumn{1}{c|}{\textbf{\begin{tabular}[c]{@{}c@{}}\textsf{DMGW}\\ \cite{Wang2019}\end{tabular}}} & \multicolumn{1}{c|}{\textbf{\begin{tabular}[c]{@{}c@{}}\textsf{DIFRNT}\\ \cite{Choi_TOG20}\end{tabular}}} & \multicolumn{1}{c|}{\textbf{\begin{tabular}[c]{@{}c@{}}\textsf{PWStableNet}\\ \cite{Zhao2020}\end{tabular}}} & \multicolumn{1}{c|}{\textbf{\begin{tabular}[c]{@{}c@{}}\textsf{GlobalFlowNet-Affine}\\   \end{tabular}}} & \textbf{\begin{tabular}[c]{@{}c@{}}\textsf{GlobalFlowNet-Full}\\ \end{tabular}} \\ \hline

\rowcolor{Gray-lighter}
\multicolumn{10}{|c|}{\textbf{Stability}   $\uparrow$} \\ \hline

\multicolumn{1}{|c|}{} &
  \multicolumn{1}{l|}{Regular} &
  \multicolumn{1}{c|}{0.781} &
  \multicolumn{1}{c|}{0.948} &
  \multicolumn{1}{c|}{0.847} &
  \multicolumn{1}{c|}{0.877} &
  \multicolumn{1}{c|}{0.795} &
  \multicolumn{1}{c|}{0.845} &
  \multicolumn{1}{c|}{0.942} &
  0.924 \\ \cline{2-10} 
\multicolumn{1}{|c|}{} &
  \multicolumn{1}{l|}{Parallax} &
  \multicolumn{1}{c|}{0.886} &
  \multicolumn{1}{c|}{0.942} &
  \multicolumn{1}{c|}{0.917} &
  \multicolumn{1}{c|}{0.914} &
  \multicolumn{1}{c|}{0.870} &
  \multicolumn{1}{c|}{0.887} &
  \multicolumn{1}{c|}{0.945} &
  0.945 \\ \cline{2-10} 
\multicolumn{1}{|c|}{} &
  \multicolumn{1}{l|}{QuickRotation} &
  \multicolumn{1}{c|}{0.949} &
  \multicolumn{1}{c|}{0.961} &
  \multicolumn{1}{c|}{0.945} &
  \multicolumn{1}{c|}{0.927} &
  \multicolumn{1}{c|}{0.898} &
  \multicolumn{1}{c|}{0.948} &
  \multicolumn{1}{c|}{0.937} &
  0.963 \\ \cline{2-10} 
\multicolumn{1}{|c|}{} &
  \multicolumn{1}{l|}{Crowd} &
  \multicolumn{1}{c|}{0.857} &
  \multicolumn{1}{c|}{0.933} &
  \multicolumn{1}{c|}{0.899} &
  \multicolumn{1}{c|}{0.898} &
  \multicolumn{1}{c|}{0.835} &
  \multicolumn{1}{c|}{0.869} &
  \multicolumn{1}{c|}{0.943} &
  0.944 \\ \cline{2-10} 
\multicolumn{1}{|c|}{} &
  \multicolumn{1}{l|}{Running} &
  \multicolumn{1}{c|}{0.784} &
  \multicolumn{1}{c|}{0.894} &
  \multicolumn{1}{c|}{0.845} &
  \multicolumn{1}{c|}{0.795} &
  \multicolumn{1}{c|}{0.757} &
  \multicolumn{1}{c|}{0.801} &
  \multicolumn{1}{c|}{0.900} &
  0.907 \\ \cline{2-10} 
\multicolumn{1}{|c|}{\multirow{-5}{*}{\textbf{NUS}}} &
  \multicolumn{1}{l|}{Zooming} &
  \multicolumn{1}{c|}{0.910} &
  \multicolumn{1}{c|}{0.950} &
  \multicolumn{1}{c|}{0.936} &
  \multicolumn{1}{c|}{0.903} &
  \multicolumn{1}{c|}{0.912} &
  \multicolumn{1}{c|}{0.855} &
  \multicolumn{1}{c|}{0.919} &
  0.919 \\ \hline
\multicolumn{1}{|c|}{\textbf{DeepStab}} &
  \multicolumn{1}{c|}{\textit{NA}} &
  \multicolumn{1}{c|}{\cellcolor[HTML]{FFFFFF}0.757} &
  \multicolumn{1}{c|}{\cellcolor[HTML]{FFFFFF}0.965} &
  \multicolumn{1}{c|}{\cellcolor[HTML]{FFFFFF}0.878} &
  \multicolumn{1}{c|}{\cellcolor[HTML]{FFFFFF}0.823} &
  \multicolumn{1}{c|}{\cellcolor[HTML]{FFFFFF}0.811} &
  \multicolumn{1}{c|}{\cellcolor[HTML]{FFFFFF}0.819} &
  \multicolumn{1}{c|}{\cellcolor[HTML]{FFFFFF}0.926} &
  \cellcolor[HTML]{FFFFFF}0.928 \\ \hline
  
  \rowcolor{Gray-lighter}
\multicolumn{10}{|c|}{\textbf{ISI}   $\uparrow$} \\ \hline
\multicolumn{1}{|c|}{} &
  \multicolumn{1}{l|}{Regular} &
  \multicolumn{1}{c|}{0.617} &
  \multicolumn{1}{c|}{0.892} &
  \multicolumn{1}{c|}{0.870} &
  \multicolumn{1}{c|}{0.812} &
  \multicolumn{1}{c|}{0.971} &
  \multicolumn{1}{c|}{0.766} &
  \multicolumn{1}{c|}{0.880} &
  0.914 \\ \cline{2-10} 
\multicolumn{1}{|c|}{} &
  \multicolumn{1}{l|}{Parallax} &
  \multicolumn{1}{c|}{0.680} &
  \multicolumn{1}{c|}{0.797} &
  \multicolumn{1}{c|}{0.802} &
  \multicolumn{1}{c|}{0.702} &
  \multicolumn{1}{c|}{0.975} &
  \multicolumn{1}{c|}{0.742} &
  \multicolumn{1}{c|}{0.816} &
  0.837 \\ \cline{2-10} 
\multicolumn{1}{|c|}{} &
  \multicolumn{1}{l|}{QuickRotation} &
  \multicolumn{1}{c|}{0.692} &
  \multicolumn{1}{c|}{0.821} &
  \multicolumn{1}{c|}{0.841} &
  \multicolumn{1}{c|}{0.715} &
  \multicolumn{1}{c|}{0.976} &
  \multicolumn{1}{c|}{0.769} &
  \multicolumn{1}{c|}{0.785} &
  0.806 \\ \cline{2-10} 
\multicolumn{1}{|c|}{} &
  \multicolumn{1}{l|}{Crowd} &
  \multicolumn{1}{c|}{0.710} &
  \multicolumn{1}{c|}{0.852} &
  \multicolumn{1}{c|}{0.848} &
  \multicolumn{1}{c|}{0.674} &
  \multicolumn{1}{c|}{0.972} &
  \multicolumn{1}{c|}{0.790} &
  \multicolumn{1}{c|}{0.849} &
  0.864 \\ \cline{2-10} 
\multicolumn{1}{|c|}{} &
  \multicolumn{1}{l|}{Running} &
  \multicolumn{1}{c|}{0.580} &
  \multicolumn{1}{c|}{0.781} &
  \multicolumn{1}{c|}{0.809} &
  \multicolumn{1}{c|}{0.656} &
  \multicolumn{1}{c|}{0.974} &
  \multicolumn{1}{c|}{0.690} &
  \multicolumn{1}{c|}{0.775} &
  0.802 \\ \cline{2-10} 
\multicolumn{1}{|c|}{\multirow{-5}{*}{\textbf{NUS}}} &
  \multicolumn{1}{l|}{Zooming} &
  \multicolumn{1}{c|}{0.656} &
  \multicolumn{1}{c|}{0.819} &
  \multicolumn{1}{c|}{0.818} &
  \multicolumn{1}{c|}{0.715} &
  \multicolumn{1}{c|}{0.973} &
  \multicolumn{1}{c|}{0.736} &
  \multicolumn{1}{c|}{0.793} &
  0.822 \\ \hline
\multicolumn{1}{|c|}{\textbf{DeepStab}} &
  \multicolumn{1}{c|}{\textit{NA}} &
  \multicolumn{1}{c|}{\cellcolor[HTML]{FFFFFF}0.681} &
  \multicolumn{1}{c|}{\cellcolor[HTML]{FFFFFF}0.942} &
  \multicolumn{1}{c|}{\cellcolor[HTML]{FFFFFF}0.887} &
  \multicolumn{1}{c|}{\cellcolor[HTML]{FFFFFF}0.858} &
  \multicolumn{1}{c|}{\cellcolor[HTML]{FFFFFF}0.776} &
  \multicolumn{1}{c|}{\cellcolor[HTML]{FFFFFF}0.800} &
  \multicolumn{1}{c|}{\cellcolor[HTML]{FFFFFF}0.880} &
  \cellcolor[HTML]{FFFFFF}0.897 \\ \hline
  
  \rowcolor{Gray-lighter}
\multicolumn{10}{|c|}{\textbf{ITF}   $\uparrow$} \\ \hline
\multicolumn{1}{|c|}{} &
  \multicolumn{1}{l|}{Regular} &
  \multicolumn{1}{c|}{18.89} &
  \multicolumn{1}{c|}{28.07} &
  \multicolumn{1}{c|}{26.95} &
  \multicolumn{1}{c|}{24.73} &
  \multicolumn{1}{c|}{22.53} &
  \multicolumn{1}{c|}{21.77} &
  \multicolumn{1}{c|}{27.35} &
  29.82 \\ \cline{2-10} 
\multicolumn{1}{|c|}{} &
  \multicolumn{1}{l|}{Parallax} &
  \multicolumn{1}{c|}{18.67} &
  \multicolumn{1}{c|}{22.16} &
  \multicolumn{1}{c|}{22.31} &
  \multicolumn{1}{c|}{20.42} &
  \multicolumn{1}{c|}{20.55} &
  \multicolumn{1}{c|}{20.38} &
  \multicolumn{1}{c|}{22.66} &
  23.44 \\ \cline{2-10} 
\multicolumn{1}{|c|}{} &
  \multicolumn{1}{l|}{QuickRotation} &
  \multicolumn{1}{c|}{19.60} &
  \multicolumn{1}{c|}{24.08} &
  \multicolumn{1}{c|}{23.20} &
  \multicolumn{1}{c|}{21.20} &
  \multicolumn{1}{c|}{21.64} &
  \multicolumn{1}{c|}{21.54} &
  \multicolumn{1}{c|}{22.46} &
  23.27 \\ \cline{2-10} 
\multicolumn{1}{|c|}{} &
  \multicolumn{1}{l|}{Crowd} &
  \multicolumn{1}{c|}{19.40} &
  \multicolumn{1}{c|}{23.54} &
  \multicolumn{1}{c|}{23.47} &
  \multicolumn{1}{c|}{19.34} &
  \multicolumn{1}{c|}{21.30} &
  \multicolumn{1}{c|}{21.30} &
  \multicolumn{1}{c|}{23.65} &
  24.34 \\ \cline{2-10} 
\multicolumn{1}{|c|}{} &
  \multicolumn{1}{l|}{Running} &
  \multicolumn{1}{c|}{17.06} &
  \multicolumn{1}{c|}{22.17} &
  \multicolumn{1}{c|}{23.28} &
  \multicolumn{1}{c|}{19.53} &
  \multicolumn{1}{c|}{19.70} &
  \multicolumn{1}{c|}{18.97} &
  \multicolumn{1}{c|}{22.27} &
  23.21 \\ \cline{2-10} 
\multicolumn{1}{|c|}{\multirow{-5}{*}{\textbf{NUS}}} &
  \multicolumn{1}{l|}{Zooming} &
  \multicolumn{1}{c|}{19.04} &
  \multicolumn{1}{c|}{23.84} &
  \multicolumn{1}{c|}{23.86} &
  \multicolumn{1}{c|}{21.34} &
  \multicolumn{1}{c|}{21.42} &
  \multicolumn{1}{c|}{20.80} &
  \multicolumn{1}{c|}{23.13} &
  24.29 \\ \hline
\multicolumn{1}{|c|}{\textbf{DeepStab}} &
  \multicolumn{1}{c|}{\textit{NA}} &
  \multicolumn{1}{c|}{\cellcolor[HTML]{FFFFFF}20.173} &
  \multicolumn{1}{c|}{\cellcolor[HTML]{FFFFFF}25.362} &
  \multicolumn{1}{c|}{\cellcolor[HTML]{FFFFFF}26.885} &
  \multicolumn{1}{c|}{\cellcolor[HTML]{FFFFFF}26.171} &
  \multicolumn{1}{c|}{\cellcolor[HTML]{FFFFFF}22.608} &
  \multicolumn{1}{c|}{\cellcolor[HTML]{FFFFFF}21.951} &
  \multicolumn{1}{c|}{\cellcolor[HTML]{FFFFFF}27.099} &
  \cellcolor[HTML]{FFFFFF}28.195 \\ \hline
  
  \rowcolor{Gray-lighter}
\multicolumn{10}{|c|}{\textbf{Crop Ratio}   $\uparrow$} \\ \hline
\multicolumn{1}{|c|}{} &
  \multicolumn{1}{l|}{Regular} &
  \multicolumn{1}{c|}{1.000} &
  \multicolumn{1}{c|}{0.834} &
  \multicolumn{1}{c|}{0.865} &
  \multicolumn{1}{c|}{0.567} &
  \multicolumn{1}{c|}{0.969} &
  \multicolumn{1}{c|}{-} &
  \multicolumn{1}{c|}{0.843} &
  0.827 \\ \cline{2-10} 
\multicolumn{1}{|c|}{} &
  \multicolumn{1}{l|}{Parallax} &
  \multicolumn{1}{c|}{1.000} &
  \multicolumn{1}{c|}{0.867} &
  \multicolumn{1}{c|}{0.780} &
  \multicolumn{1}{c|}{0.438} &
  \multicolumn{1}{c|}{0.970} &
  \multicolumn{1}{c|}{-} &
  \multicolumn{1}{c|}{0.838} &
  0.814 \\ \cline{2-10} 
\multicolumn{1}{|c|}{} &
  \multicolumn{1}{l|}{QuickRotation} &
  \multicolumn{1}{c|}{1.000} &
  \multicolumn{1}{c|}{0.845} &
  \multicolumn{1}{c|}{0.555} &
  \multicolumn{1}{c|}{0.453} &
  \multicolumn{1}{c|}{0.658} &
  \multicolumn{1}{c|}{-} &
  \multicolumn{1}{c|}{0.835} &
  0.808 \\ \cline{2-10} 
\multicolumn{1}{|c|}{} &
  \multicolumn{1}{l|}{Crowd} &
  \multicolumn{1}{c|}{1.000} &
  \multicolumn{1}{c|}{0.853} &
  \multicolumn{1}{c|}{0.825} &
  \multicolumn{1}{c|}{0.483} &
  \multicolumn{1}{c|}{0.962} &
  \multicolumn{1}{c|}{-} &
  \multicolumn{1}{c|}{0.831} &
  0.802 \\ \cline{2-10} 
\multicolumn{1}{|c|}{} &
  \multicolumn{1}{l|}{Running} &
  \multicolumn{1}{c|}{1.000} &
  \multicolumn{1}{c|}{0.826} &
  \multicolumn{1}{c|}{0.701} &
  \multicolumn{1}{c|}{0.476} &
  \multicolumn{1}{c|}{0.942} &
  \multicolumn{1}{c|}{-} &
  \multicolumn{1}{c|}{0.799} &
  0.767 \\ \cline{2-10} 
\multicolumn{1}{|c|}{\multirow{-5}{*}{\textbf{NUS}}} &
  \multicolumn{1}{l|}{Zooming} &
  \multicolumn{1}{c|}{1.000} &
  \multicolumn{1}{c|}{0.768} &
  \multicolumn{1}{c|}{0.711} &
  \multicolumn{1}{c|}{0.513} &
  \multicolumn{1}{c|}{0.903} &
  \multicolumn{1}{c|}{-} &
  \multicolumn{1}{c|}{0.834} &
  0.791 \\ \hline
\multicolumn{1}{|c|}{\textbf{DeepStab}} &
  \multicolumn{1}{c|}{\textit{NA}} &
  \multicolumn{1}{c|}{\cellcolor[HTML]{FFFFFF}1.000} &
  \multicolumn{1}{c|}{\cellcolor[HTML]{FFFFFF}-} &
  \multicolumn{1}{c|}{\cellcolor[HTML]{FFFFFF}0.791} &
  \multicolumn{1}{c|}{\cellcolor[HTML]{FFFFFF}0.471} &
  \multicolumn{1}{c|}{\cellcolor[HTML]{FFFFFF}0.972} &
  \multicolumn{1}{c|}{\cellcolor[HTML]{FFFFFF}-} &
  \multicolumn{1}{c|}{\cellcolor[HTML]{FFFFFF}0.821} &
  \cellcolor[HTML]{FFFFFF}0.790 \\ \hline
  
  \rowcolor{Gray-lighter}
\multicolumn{10}{|c|}{\textbf{AGDMR}   $\uparrow$} \\ \hline
\multicolumn{1}{|c|}{} &
  \multicolumn{1}{l|}{Regular} &
  \multicolumn{1}{c|}{0.000} &
  \multicolumn{1}{c|}{0.812} &
  \multicolumn{1}{c|}{0.592} &
  \multicolumn{1}{c|}{-0.384} &
  \multicolumn{1}{c|}{0.305} &
  \multicolumn{1}{c|}{0.411} &
  \multicolumn{1}{c|}{0.735} &
  0.821 \\ \cline{2-10} 
\multicolumn{1}{|c|}{} &
  \multicolumn{1}{l|}{Parallax} &
  \multicolumn{1}{c|}{0.000} &
  \multicolumn{1}{c|}{0.439} &
  \multicolumn{1}{c|}{0.312} &
  \multicolumn{1}{c|}{-3.354} &
  \multicolumn{1}{c|}{0.234} &
  \multicolumn{1}{c|}{0.171} &
  \multicolumn{1}{c|}{0.573} &
  0.635 \\ \cline{2-10} 
\multicolumn{1}{|c|}{} &
  \multicolumn{1}{l|}{QuickRotation} &
  \multicolumn{1}{c|}{0.000} &
  \multicolumn{1}{c|}{0.087} &
  \multicolumn{1}{c|}{-2.360} &
  \multicolumn{1}{c|}{-6.685} &
  \multicolumn{1}{c|}{-2.069} &
  \multicolumn{1}{c|}{-0.398} &
  \multicolumn{1}{c|}{0.297} &
  0.296 \\ \cline{2-10} 
\multicolumn{1}{|c|}{} &
  \multicolumn{1}{l|}{Crowd} &
  \multicolumn{1}{c|}{0.000} &
  \multicolumn{1}{c|}{0.402} &
  \multicolumn{1}{c|}{0.283} &
  \multicolumn{1}{c|}{-3.518} &
  \multicolumn{1}{c|}{0.265} &
  \multicolumn{1}{c|}{0.232} &
  \multicolumn{1}{c|}{0.567} &
  0.632 \\ \cline{2-10} 
\multicolumn{1}{|c|}{} &
  \multicolumn{1}{l|}{Running} &
  \multicolumn{1}{c|}{0.000} &
  \multicolumn{1}{c|}{0.687} &
  \multicolumn{1}{c|}{0.598} &
  \multicolumn{1}{c|}{-0.709} &
  \multicolumn{1}{c|}{0.446} &
  \multicolumn{1}{c|}{0.225} &
  \multicolumn{1}{c|}{0.684} &
  0.742 \\ \cline{2-10} 
\multicolumn{1}{|c|}{\multirow{-5}{*}{\textbf{NUS}}} &
  \multicolumn{1}{l|}{Zooming} &
  \multicolumn{1}{c|}{0.000} &
  \multicolumn{1}{c|}{0.680} &
  \multicolumn{1}{c|}{0.338} &
  \multicolumn{1}{c|}{-1.165} &
  \multicolumn{1}{c|}{0.160} &
  \multicolumn{1}{c|}{0.178} &
  \multicolumn{1}{c|}{0.625} &
  0.686 \\ \hline
\multicolumn{1}{|c|}{\textbf{DeepStab}} &
  \multicolumn{1}{c|}{\textit{NA}} &
  \multicolumn{1}{c|}{0.000} &
  \multicolumn{1}{c|}{0.647} &
  \multicolumn{1}{c|}{0.554} &
  \multicolumn{1}{c|}{-0.019} &
  \multicolumn{1}{c|}{0.278} &
  \multicolumn{1}{c|}{0.486} &
  \multicolumn{1}{c|}{0.786} &
  0.815 \\ \hline
  
  \rowcolor{Gray-lighter}
\multicolumn{10}{|c|}{\textbf{Distortion}   $\uparrow$} \\ \hline
\multicolumn{1}{|c|}{} &
  \multicolumn{1}{l|}{Regular} &
  \multicolumn{1}{c|}{1.000} &
  \multicolumn{1}{c|}{0.970} &
  \multicolumn{1}{c|}{0.947} &
  \multicolumn{1}{c|}{0.797} &
  \multicolumn{1}{c|}{0.980} &
  \multicolumn{1}{c|}{0.979} &
  \multicolumn{1}{c|}{0.997} &
  0.978 \\ \cline{2-10} 
\multicolumn{1}{|c|}{} &
  \multicolumn{1}{l|}{Parallax} &
  \multicolumn{1}{c|}{1.000} &
  \multicolumn{1}{c|}{0.908} &
  \multicolumn{1}{c|}{0.817} &
  \multicolumn{1}{c|}{0.398} &
  \multicolumn{1}{c|}{0.885} &
  \multicolumn{1}{c|}{0.866} &
  \multicolumn{1}{c|}{0.997} &
  0.978 \\ \cline{2-10} 
\multicolumn{1}{|c|}{} &
  \multicolumn{1}{l|}{QuickRotation} &
  \multicolumn{1}{c|}{1.000} &
  \multicolumn{1}{c|}{0.849} &
  \multicolumn{1}{c|}{0.482} &
  \multicolumn{1}{c|}{0.040} &
  \multicolumn{1}{c|}{0.889} &
  \multicolumn{1}{c|}{0.936} &
  \multicolumn{1}{c|}{0.936} &
  0.887 \\ \cline{2-10} 
\multicolumn{1}{|c|}{} &
  \multicolumn{1}{l|}{Crowd} &
  \multicolumn{1}{c|}{1.000} &
  \multicolumn{1}{c|}{0.902} &
  \multicolumn{1}{c|}{0.931} &
  \multicolumn{1}{c|}{0.019} &
  \multicolumn{1}{c|}{0.975} &
  \multicolumn{1}{c|}{0.976} &
  \multicolumn{1}{c|}{0.997} &
  0.972 \\ \cline{2-10} 
\multicolumn{1}{|c|}{} &
  \multicolumn{1}{l|}{Running} &
  \multicolumn{1}{c|}{1.000} &
  \multicolumn{1}{c|}{0.886} &
  \multicolumn{1}{c|}{0.908} &
  \multicolumn{1}{c|}{0.057} &
  \multicolumn{1}{c|}{0.968} &
  \multicolumn{1}{c|}{0.967} &
  \multicolumn{1}{c|}{0.997} &
  0.961 \\ \cline{2-10} 
\multicolumn{1}{|c|}{\multirow{-5}{*}{\textbf{NUS}}} &
  \multicolumn{1}{l|}{Zooming} &
  \multicolumn{1}{c|}{1.000} &
  \multicolumn{1}{c|}{0.891} &
  \multicolumn{1}{c|}{0.834} &
  \multicolumn{1}{c|}{0.343} &
  \multicolumn{1}{c|}{0.971} &
  \multicolumn{1}{c|}{0.969} &
  \multicolumn{1}{c|}{0.993} &
  0.898 \\ \hline
\multicolumn{1}{|c|}{\textbf{DeepStab}} &
  \multicolumn{1}{c|}{\textit{NA}} &
  \multicolumn{1}{c|}{\cellcolor[HTML]{FFFFFF}0} &
  \multicolumn{1}{c|}{\cellcolor[HTML]{FFFFFF}0.647} &
  \multicolumn{1}{c|}{\cellcolor[HTML]{FFFFFF}0.554} &
  \multicolumn{1}{c|}{\cellcolor[HTML]{FFFFFF}-0.019} &
  \multicolumn{1}{c|}{\cellcolor[HTML]{FFFFFF}0.966} &
  \multicolumn{1}{c|}{\cellcolor[HTML]{FFFFFF}0.486} &
  \multicolumn{1}{c|}{\cellcolor[HTML]{FFFFFF}0.786} &
  \cellcolor[HTML]{FFFFFF}0.81 \\ \hline
\end{tabular}%
}
\caption{Numerical performance comparison for different video stabilization methods. All measures are averaged across 142 videos from \cite{videostabLiu} and 60 videos from \cite{DeepStab2018}. Higher values are better for all measures. \textit{Note that the entries for Crop Ratios are left blank for methods without publicly available video results or the source code  did not implement cropping invalid region}}
\label{tab:numerical_perf}
\end{table*}
\label{sec:experiments}
\noindent\textbf{Training Details}: For global motion estimation, \textsc{GlobalFlowNet} was trained on randomly chosen consecutive frame pairs from 10K videos in the RealEstate10K dataset \cite{Zhou2018_IJCV}. The pre-trained PWC-Net model from \cite{Sun2018} was used for the teacher network. To train the student, we chose a batch-size of 16 and 200K iterations with the Adam optimizer. The training time was two days with NVIDIA Quadro RTX 5000 16GB GDDR6 Graphics Card and Intel Xeon E5-2620 CPU. Unlike \cite{Yu2020}, we do not perform stabilization, but only motion estimation, via a neural network. Hence no synthetic distortions need to be introduced in the dataset for training. 

\noindent\textbf{Dataset, Parameters and Comparisons:} We now present experimental results to validate our approach on a total of 202 videos, i.e., on \emph{all}(142) videos belonging to the following categories from the dataset \cite{videostabLiu}: regular, large parallax, quick rotation, crowd, running, zooming, as well all 60 videos from the Deepstab dataset \cite{DeepStab2018}. For our algorithm, 
%the following parameters were set: crop ratio lower bound $\kappa \triangleq 0.85$ in Sec.~\ref{subsec:globalStabilization}, 
we set $W_R \triangleq 16$ in Sec.~\ref{subsec:smooth_dct}. Our algorithm was compared against the following recent state-of-the-art techniques: (1) the `bundled camera paths' (\textsf{BCP}) approach from \cite{Liu2013}, (2) the `learning video stabilization' (\textsf{LVS}) approach from \cite{Yu2020} (which is shown to be faster and superior to the earlier approach in \cite{Yu2019}), (3) the deep multi-grid warping (\textsf{DMGW}) approach from \cite{Wang2019}, (4) the neural network approach called \textsf{PWStableNet} from \cite{Zhao2020}, and (5) the frame interpolation approach (\textsf{DIFRNT}) from \cite{Choi2020}. For our approach, we compare the outputs from the affine-only stage (steps 1--4 from Alg.~\ref{algo:videostab}) to those from the complete algorithm. We term these as \textsf{GlobalFlowNet-Affine} and \textsf{GlobalFlowNet-Full} respectively. For all competing methods, we used the implementations provided by the authors with their recommended parameters. For \textsf{BCP}, no code was available, but we compared with the sample results provided by the authors.

\noindent\textbf{Visual comparison:} Visual results for global motion estimation can be found in Fig.~\ref{fig:theory} and Figs. 2 and 3 of \cite{suppmat}. The visual outputs of the different methods can be observed in the supplemental material \cite{suppmat} on different videos from each of the six aforementioned categories. These results reveal the visual stability of \textsf{GlobalFlowNet-Affine} and \textsf{GlobalFlowNet-Full}, as compared to competing methods. 

\noindent\textbf{Numerical comparison:} Subjective visual quality apart, we objectively compared the competing algorithms in terms of the following quality measures, results for which are presented in Table \ref{tab:numerical_perf}. 
\underline{\texttt{1. Stability}} \cite{Liu2013}: This measure is computed by determining the frame-to-frame translation and rotation parameters in the stabilized videos. These parameter sequences (across time) are then reconstructed using only the first $K_F \triangleq 6$ DFT coefficients (other than DC). The stability measure is equal to the minimum (over the four parameters) of the ratios of the $\ell_2$ norm of the reconstructed sequence to that of the original sequence. The intuition is based on the smoothness (dominant low frequency content) of these parameter sequences in stable videos.
\underline{\texttt{2. Distortion \cite{Liu2013}}}: This measure is computed from the average (across the frames) ratio of the smaller to the larger eigenvalue of the affine transformation matrix between the corresponding frames of the unstable and stabilized videos. A larger value (closer to 1) is desirable. This measure has some limitations as it will falsely yield an optimal value when the supposedly stabilized video is just a copy of the original unstable video (as the affine transformation between two identical frames is identity). We argue that other measures such as \texttt{ISI}, \texttt{ITF} and our new measure \texttt{AGMDR} introduced in the main paper, are more appropriate quality measures. But we are including comparisons using \texttt{Distortion} here, owing to its wide usage in video stabilization.
\underline{3. Inter-frame Similarity Index (\texttt{ISI})} \cite{Guilluy2018}: This is the average inter-frame SSIM \cite{Wang2009}, expressed as $\frac{1}{T-1} \sum_{i=1}^{T-1} \textrm{SSIM}(I_i,I_{i+1})$. The intuition is that consecutive frames of a stabilized video would have higher pairwise SSIM than in an unstable one. 
\underline{4. Inter-frame Transformation Fidelity} (\texttt{ITF}) \cite{Guilluy2018}: This is the average PSNR between consecutive frames and is expressed as $\frac{1}{T-1} \sum_{i=1}^{T-1} \textrm{PSNR}(I_i,I_{i+1})$. %The intuition is that consecutive frames of a stabilized video would have higher pairwise PSNR than in an unstable one. 
\underline{\texttt{5. Crop ratio} \cite{Liu2013}} defined earlier in Sec.~\ref{subsec:globalStabilization}.
\underline{6. Average Global Motion Difference Ratio (\texttt{AGMDR})}: This is a new measure which we propose here. It is equal to one minus the ratio of the total magnitude of the difference in the global motion between consecutive pairs of frames in the stabilized video to that in the unstable video. It is computed as $1-\dfrac{\sum_{i=2}^{T-1} \|\boldsymbol{f^s_S}^{(i,i+1)}-\boldsymbol{f^s_S}^{(i-1,i)}\|_2}{\sum_{i=2}^{T-1}\|\boldsymbol{f^u_S}^{(i,i+1)}-\boldsymbol{f^u_S}^{(i-1,i)}\|_2}$. Here $\boldsymbol{f^s_S}, \boldsymbol{f^u_S}$ denote the inter-frame global motion in the stabilized and unstable videos respectively (unaffected by moving objects) and are computed using \textsc{GlobalFlowNet} from Sec.~\ref{subsec:low_pass_module}.  

The values of all these measures except \texttt{ITF} lie between 0 to 1. Higher values for all these measures indicate better performance. The first three measures, though widely used in video stabilization \cite{Yu2019,Yu2020,Liu2013,PLiu2019,Zhao2020}, are based entirely on affine motion approximation, which as argued earlier, is an inaccurate motion model. On the other hand, \texttt{AGMDR} is based on a more general motion model. The intuition behind it is that the global motion across consecutive frames in a stable video should vary smoothly. Also, \texttt{AGMDR} is computationally very efficient, and by construction resilient against moving objects.

%In \cite{suppmat}, we also show results on \texttt{AGMR}, a related measure based on relative optical flow reduction, as well as two related measures based on salient feature points.  

\noindent\textbf{Discussion on results:} Due to the absence of a single universally accepted evaluation measure for video stabilization, a holistic view of different needs/measures is required to draw a conclusion from numerical scores. On scores such as \texttt{Stability}, \texttt{ISI}, \texttt{ITF} and \texttt{AGMDR}, our techniques outperform the competing techniques \textsf{LVS},  \textsf{BCP}, \textsf{DMGW} and \textsf{PWStableNet} in most of the video categories, as can be seen in Table \ref{tab:numerical_perf}. Our methods outperform \textsf{DIFRNT} on all measures except \texttt{ISI}. However, we would like to note that, \textsf{DIFRNT}  being based on iterative frame interpolation, would by design repeatedly reinforce similarity between adjacent frames over iterations, thus producing higher \texttt{ISI}.  For \texttt{AGMDR}, some stabilization techniques produced jerkier optical flow in a few frames as compared to the original unstable video, leading to negative values of this measure. Our method produces less geometric/visual distortion than other methods as also evidenced by better \texttt{ISI}, \texttt{ITF} and \texttt{Distortion} scores (also see \cite{suppmat}). Also, the video `Ablation.mp4' in \cite{suppmat} clearly shows the advantages of \textsc{GlobalFlowNet-Full} over \textsc{GlobalFlowNet-Affine}.
\vspace{-0.2in}
\section{Conclusion}
\label{sec:conclusion}
We have presented a novel and intuitive video stabilization technique which uses a teacher-student network to obtain frame-to-frame global motion expressed by a small set of transform coefficients. We present novel ways to smooth the MPS in order to generate a stabilized video. Our method is quite general in nature and can potentially be extended using alternative architectures for optical flow (other than PWC-Net) and alternative transforms for motion representation (other than DCT). Our technique also inspired us to propose a novel quality measure to evaluate video stabilization. Moreover, the presented algorithm is computationally efficient and typically requires only $\sim$0.06 seconds and $\sim$0.5 seconds per frame (of size $480 \times 640$) for \textsf{GlobalFlowNet-Affine} and \textsf{GlobalFlowNet-Full} respectively.

\noindent The second stage \textsf{GlobalFlowNet-Full} of our algorithm will produce sub-optimal results if the area of a \emph{single dominant} foreground object is large and comparable to that of the background. Note that in general, multiple independently moving small foregrounds do not pose a problem as their motion will generally be very different from each other and their combined motion will not dominate background) -- see \cite[Sec. 6]{suppmat}. In case of a single large foreground, the motion estimates will be biased towards either the foreground or the background, producing motion artifacts. The first stage \textsf{GlobalFlowNet-Affine} is more resilient to such situations. However a principled method to handle this in the second stage is left to future work. The cropping ratio can be further improved by adding \cite{Xu_2021_ICCV}. The motion estimation can be improved using \cite{Hruby_2022_CVPR}. These are avenues for future work, as also extending our algorithm on rolling shutter videos and videos with significant motion blur.% Also, we have not tested our algorithm on rolling shutter videos or those with significant motion blur. These are avenues for future work, as also applications of video stabilization in drone photography, capsule endoscopy and underwater imaging. 

{\small
\bibliographystyle{ieee_fullname}
\bibliography{main}
}

\end{document}